\pdfoutput=1
\documentclass[11pt,a4paper]{article}
\usepackage[nohyperref]{naaclhlt2018}
\usepackage{times}
\usepackage{latexsym}
\usepackage{tikz}
\usepackage{pgfplots}
\usepackage{subfig}
\usepackage{numprint}
\usepackage{multicol}
\usepackage{multirow}
\usepackage[tbtags]{amsmath}
\usepackage{amssymb}
\usepackage{comment}
\usepackage{booktabs}
\usepackage{graphicx}
\usepackage{relsize}
\usepackage{booktabs}
\usepackage{tkz-graph}
\usepackage[htt]{hyphenat}
\usepackage{bbding}
\usepackage{xcolor}
\usepackage{relsize}
\usepackage{lipsum}
\usepackage{url}
\usepackage{tkz-euclide}
\usetkzobj{all}
\usetikzlibrary{shapes,decorations}
\aclfinalcopy % Uncomment this line for the final submission
\title{What do you mean, \textsc{bert}? \\ Assessing \textsc{bert} as a Distributional Semantics Model}

\author{Timothee Mickus \\
  Universit\'e de Lorraine \\
  CNRS, ATILF \\ 
  {\tt tmickus@atilf.fr} \\\And
  Denis Paperno \\
  Utrecht University \\
  {\tt d.paperno@uu.nl} \\\And
  Mathieu Constant \\
  Universit\'e de Lorraine \\
  CNRS, ATILF \\ 
  {\tt mconstant@atilf.fr} \\\And
  Kees van Deemter \\
  Utrecht University \\
  {\tt c.j.vandeemter@uu.nl} \\}

\date{}

\definecolor{thebluebase}{rgb}{.2,.2,.7}
\colorlet{theblue}{thebluebase!85!white}
\definecolor{thegreen}{rgb}{.53, .66, .42}

\begin{document}
\maketitle
\begin{abstract}

Contextualized word embeddings, i.e.\ vector representations for words in context, are naturally seen as an extension of previous noncontextual distributional semantic models.
In this work, we focus on \textsc{bert}, a deep neural network that produces contextualized embeddings and has set the state-of-the-art in several semantic tasks, and study the semantic coherence of its embedding space.
While showing a tendency towards coherence, \textsc{bert} does not fully live up to the natural expectations for a semantic vector space. In particular, we find that the position of the sentence in which a word occurs, while having no meaning correlates, leaves a noticeable trace on the word embeddings and disturbs similarity relationships.

\end{abstract}

\section{Introduction}

A recent success story of \textsc{nlp}, \textsc{bert} \citep{Devlin18Bert} stands at the crossroad of two key innovations that have brought about significant improvements over previous state-of-the-art results. On the one hand, \textsc{bert} models are an instance of contextual embeddings \citep{McCann17CoVe,Peters18ELMo}, which have been shown to be subtle and accurate representations of words within sentences. On the other hand, \textsc{bert} is a  variant of the Transformer architecture \citep{Vaswani17} which has set a new state-of-the-art on a wide variety of tasks ranging from machine translation \citep{ott18scalingNMT} to language modeling \citep{dai19tfxl}. %As a testament of their efficiency, 
\textsc{bert}-based models have significantly increased state-of-the-art over the \textsc{glue} benchmark for natural language understanding \citep{wang2019glue} and most of the best scoring models for this benchmark include or elaborate on \textsc{bert}. Using \textsc{bert} representations has become in many cases a new standard approach: for instance, all submissions at the recent shared task on gendered pronoun resolution \citep{webster-etal-2019-gendered} were based on \textsc{bert}. Furthermore, \textsc{bert} serves both as a strong baseline and as a basis for a fine-tuned state-of-the-art word sense disambiguation pipeline \citep{superglue}.

%Remarkably, although a clear parallel is well established between `traditional' non-contextual embeddings and the theory of distributional semantics \citep[e.g.]{lenci2018distributional,Boleda2019DSandLT}, the same has not been held for contextual embeddings: for instance \citet{Westera19DSandEntailment} explicitly considers only `context-invariant' representations as distributional semantics. This discrepancy may be attributed to the original remark by \citet{Peters18ELMo} that contextual embeddings are ``functions of the entire input sentence''---whereas traditional embeddings map words to vectors.
%
%To address this lack, we here propose to study whether contextual embeddings can be understood as a particular case of distributional semantics. We focus our investigations on the characteristic trait of distributional semantics models (\textsc{dsm}) that similar objects have similar representations. More specifically, we investigate whether \textsc{bert} embeddings describe a coherent latent semantic space: can we coherently associate word meanings with regions of the vector space? 

%A trend of research concerns itself with understanding the mechanical behavior of specific architectures; in the case of Transformers in general, and \textsc{bert} in particular, such 
Analyses aiming to understand the mechanical behavior of Transformers in general, and \textsc{bert} in particular, have suggested that they compute word representations through implicitly learned syntactic operations \citep[a.o.]{raganato-tiedemann-2018-analysis,Clark19BertLook,Coenen19BERTGeometry,jawahar19BERTStructLang}: representations computed through the `attention' mechanisms of Transformers can arguably be seen as weighted sums of intermediary representations from the previous layer, %and it has been shown that
with many attention heads assigning higher weights to syntactically related tokens \citep[however, contrast with ][]{Brunner2019Validity,serrano2019Interpretable}. %Cf.\ also \citet{Linzen2016AssessingLSTMSyntax,lakretz19GMcells} for related studies applied to other neural architectures. % when computing the intermediary representation for a given token.

%More generally, another type of analyses evaluates the adequacy of specific models with a given theory---e.g. linguistic or neurological ---and
%
Complementing these previous studies, in this paper we adopt a more theory-driven lexical semantic perspective. While a clear parallel was established between `traditional' noncontextual embeddings and the theory of distributional semantics \citep[a.o.][]{lenci2018distributional,Boleda2019DSandLT}, this link is not automatically extended to contextual embeddings: some authors \citep{Westera19DSandEntailment} even explicitly consider only ``context-invariant'' representations as distributional semantics. 
Hence we study to what extent \textsc{bert}, as a contextual embedding architecture, satisfies the properties expected from a natural contextualized extension of distributional semantics models (\textsc{dsm}s). 

\textsc{dsm}s assume that meaning is derived from use in context. \textsc{dsm}s are nowadays systematically represented using vector spaces \citep{lenci2018distributional}. They generally map each word in the domain of the model to a numeric vector on the basis of distributional criteria; vector components are inferred from text data. \textsc{dsm}s have also been computed for linguistic items other than words, e.g., word senses---based both on meaning inventories \citep{Rothe15synsets} and word sense induction techniques \citep{Bartunov15Adagram}---or meaning exemplars \citep{Reisinger2010multiproto,Erk2010ExemplarBasedMF,Reddy2011DynamicAS}.
The default approach has however been to produce representations for word types. Word properties encoded by \textsc{dsm}s vary from morphological information \citep{Marelli2015,Bonamisubmitted} to geographic information \citep{Louwerse2009}, to social stereotypes \citep{Bolukbasi2016} and to referential properties \citep{Herbelot2015}.

A reason why contextualized embeddings have not been equated to distributional semantics may lie in that they are ``functions of the entire input sentence'' \citep{Peters18ELMo}.
%Contextualized embeddings %such as \textsc{bert} 
%have introduced some key differences regarding what is to be encoded in a vector representation: 
Whereas traditional \textsc{dsm}s match word \textsl{types} with numeric vectors, contextualized embeddings produce distinct vectors per \textsl{token}. 
Ideally, the contextualized nature of these embeddings should reflect the semantic nuances that context induces in the meaning of a word---with varying degrees of subtlety, ranging from broad word-sense disambiguation (e.g.\ `bank' as a river embankment or as a financial institution) to narrower subtypes of word usage (`bank' as a corporation or as a physical building) and to more context-specific nuances.

Regardless of how apt contextual embeddings such as \textsc{bert} are at capturing increasingly finer semantic distinctions, we expect the contextual variation to preserve the basic \textsc{dsm} properties. Namely, we expect that the space structure encodes meaning similarity and that variation within the embedding space is semantic in nature.  %establishing a parallel between them and distributional semantics requires that basic properties of \textsc{dsm}s be observed in embeddings.
%The property we study in particular is that
Similar words should be represented with similar vectors, and only semantically pertinent distinctions should affect these representations. 
We connect our study with previous work in section \ref{sec:theory} before detailing the two approaches we followed. First, we verify in section \ref{sec:word-type-cohesion} that \textsc{bert} embeddings form natural clusters when grouped by word types, which on any account should be groups of similar words and thus be assigned similar vectors. Second, we test in sections \ref{sec:cross-sentence-coherence-per-type} and \ref{sec:lexical-contrasts-and-cross-sentence-coherence} that contextualized word vectors do not encode semantically irrelevant features: in particular, leveraging some knowledge from the architectural design of \textsc{bert}, we address whether there is no systematic difference between \textsc{bert} representations in odd and even sentences of running text---a property we refer to as \textsl{cross-sentence coherence}. In section \ref{sec:cross-sentence-coherence-per-type}, we test whether we can observe cross-sentence coherence for single tokens, whereas in section \ref{sec:lexical-contrasts-and-cross-sentence-coherence} we study to what extent incoherence of representations across sentences affects the similarity structure of the semantic space. We summarize our findings in section \ref{sec:ccl}.

\section{Theoretical background \& connections} \label{sec:theory}

%In this section, we recount earlier connections of word embeddings and distributional semantics and stress why recent analyses of \textsc{bert} cannot be reemployed to establish such a connection. We also study what consequences arise from the fact that \textsc{bert} embeddings are ``functions of the entire input sentence'' \citep{Peters18ELMo}.

%\subsection{Distributional semantics} 

%Word embeddings have proven to be a key element to most \textsc{nlp} architectures and in particular to neural networks, where they have found a wide array of application ranging from building models of cognition \citep{Landauer1997,Marelli2015} to predicting semantic similarity judgments \citep{agirre2009study} and to initializing neural machine translation systems \citep{artetxe2017unsupervised,lample2017unsupervised}. Ever since the seminal work of \citet{Bengio2003}, neural approaches to \textsc{nlp} have been representing linguistic units as vectors in high-dimensional spaces.
   
%Most importantly, 
Word embeddings have been said to be `all-purpose' representations, capable of unifying the otherwise heterogeneous domain that is \textsc{nlp} \citep{Turney2010}. To some extent this claim spearheaded the evolution of \textsc{nlp}: focus recently shifted from  task-specific architectures with limited applicability to universal architectures requiring little to no adaptation \citep[a.o.]{Radford2018,Devlin18Bert,Radford2019,xlnet,liu19roberta}.

Word embeddings are linked to the distributional hypothesis, according to which ``you shall know a word from the company it keeps'' \citep{Firth1958}. Accordingly, the meaning of a word can be inferred from the effects it has on its context \citep{Harris54}; %It has therefore largely been assimilated to non-fregean semantics: whereas \citet{Frege1892} held that the meaning of a composed expression could be determined using the meaning of its components and the rules used to compose them, \citet{Quine1960} stressed that the exact meaning of a sentence could not be derived without understanding the language of said sentence in its entirety.
as this framework equates the meaning of a word to the set of its possible usage contexts, it corresponds more to holistic theories of meaning \citep[a.o.]{Quine1960} than to truth-value accounts \citep[a.o.]{Frege1892}.
In early works on word embeddings \citep[e.g.]{Bengio2003}, a straightforward parallel between word embeddings and distributional semantics could be made: the former are distributed representations of word meaning, the latter a theory stating that a word's meaning is drawn from its distribution. In short, word embeddings could be understood as a vector-based implementation of the distributional hypothesis. This parallel is much less obvious for contextual embeddings: are constantly changing representations truly an apt description of the meaning of a word?

More precisely, the literature on distributional semantics has put forth and discussed many mathematical properties of embeddings: embeddings are equivalent to count-based matrices \citep{Levy2014}, expected to be linearly dependant \citep{Arora16LinearWordsenses}, expressible as a unitary matrix \citep{Smith17Offline} or as a perturbation of an identity matrix \citep{Yin28Dimensionality}. All these properties have however been formalized for non-contextual embeddings: they were formulated using the tools of matrix algebra, under the assumption that embedding matrix rows correspond to word types. Hence they cannot be applied as such to contextual embeddings. %Likewise, many remarks previously made on word embeddings---e.g. the equivalence between neural embeddings and count-based matrices \citep{Levy2014}---do not hold. %In all, the formulation of many previous studies on distributional semantics and word embeddings assume word vectors per type rather than per tokens and thus cannot be applied as is to \textsc{bert}.
This disconnect in the literature %between the properties of context-invariant representations and their extension to contextualized embeddings 
leaves unanswered the question of what consequences there are to framing contextualized embeddings as \textsc{dsm}s.

%As a matter of fact, 
The analyses that contextual embeddings have been subjected to differ from most analyses of distributional semantics models. %, and are generally more focused on the mechanics of the architecture or its performances on downstream applications.
\citet{Peters18ELMo} analyzed through an extensive ablation study of \textsc{elm}o what information is captured by each layer of their architecture. \citet{Devlin18Bert} discussed what part of their architecture is critical to the performances of \textsc{bert}, comparing pre-training objectives, number of layers and training duration. 
Other works \citep{raganato-tiedemann-2018-analysis,Hewitt2019ASP,Clark19BertLook,voita19pruninghead,Michel19Sixteen} have introduced specific procedures to understand how attention-based architectures function on a mechanical level. %, %: the most promising of these being `Layer-wise Relevance Propagation' \citep{bach-plos15,ding-etal-2017-visualizing,voita19pruninghead}.
%generally focusing on either attention matrices or representations across layers. 
Recent research has however questioned the pertinence of these attention-based analyses \citep{serrano2019Interpretable,Brunner2019Validity}; moreover %in any event
these works have focused more on the inner workings of the networks than on their adequacy with theories of meaning.
%
%fact that the motivations driving these works differ from ours, we see that---although they provide promising clues and facts to account for---most of these analyses cannot be reemployed as such if we want to assess whether \textsc{bert} contextual embeddings depict a coherent semantic space on their own.
%First, for a fair comparison of \textsc{bert} with previous embedding architectures, we must restrict ourselves to a study of the output contextual embeddings of \textsc{bert} itself, and stay away from considerations on its inner workings. 
%\footnote
%Many also involve a learned classifier to `extract' information from the embeddings, yet this methodology may conflict with the intended purpose of studying the representations themselves \citep{Wieting2019randsent,cover65}. 
%

One trait of \textsc{dsm}s that is very often encountered, discussed and exploited in the literature is the fact that the relative positions of embeddings are not random. Early vector space models, by design, required that word with similar meanings lie near one another \citep{Salton1975VSM}; as a consequence, regions of the vectors space describe coherent semantic fields.\footnote{Vectors encoding contrasts between words are also expected to be coherent \citep{Mikolov2013a}, although this assumption has been subjected to criticism \citep{linzen2016issues}.}
Despite the importance of this characteristic, the question whether \textsc{bert} contextual embeddings depict a coherent semantic space on their own has been left mostly untouched by papers focusing on analyzing \textsc{bert} or Transformers \citep[with some exceptions, e.g.][]{Coenen19BERTGeometry}. Moreover, many analyses of how meaning is represented in attention-based networks or contextual embeddings  include ``probes'' (learned models such as classifiers%, \textsc{svm}s or \textsc{mlp}s
) as part of their evaluation setup to `extract' information from the embeddings \citep[e.g.]{Peters18ELMo,tang-etal-2018-wsd,Coenen19BERTGeometry,chang2019means}. Yet this methodology has been criticized as potentially conflicting with the intended purpose of studying the representations themselves \citep{Wieting2019randsent,cover65}; cf.\ also \citet{Hewitt2019ProbingProbes} for a discussion. We refrain from using learned probes in favor of a more direct assessment of the coherence of the semantic space.
% 

%\section{word type clusters}
\section{Experiment 1: Word Type Cohesion}  
\label{sec:word-type-cohesion}

%We begin by % A natural follow-up on our previous experiments is to 
%assessing whether contextualized representations like that of \textsc{bert} depict a coherent semantic space, so as to establish a stronger connection with distributional semantics. %: the extent to which we can expect \textsc{rvd} is a fit function for semantic composition is bounded by the extent to which the word embeddings on which we compute \textsc{rvd} behave like a semantic space.
%We set aside mechanical questions pertaining to how \textsc{bert} functions and instead study the behavior of these contextual embeddings externally, effectively treating the algorithm as a black box most of the time. In a sense, the focus of our experiments is on conceptual simplicity.
The trait of distributional spaces that we focus on in this study is that similar words should lie in similar regions of the semantic space. This should hold all the more so for identical words, which ought to be be maximally similar. By design, contextualized embeddings like \textsc{bert} exhibit variation within vectors corresponding to identical word types. Thus, if \textsc{bert} is a \textsc{dsm}, we expect that word types form natural, distinctive clusters %of embeddings occupying distinctive regions 
in the embedding space.
Here, we assess the coherence of word type clusters by means of their \textsl{silhouette scores} \citep{Rousseeuw1987silhouettes}.

\subsection{Data \& Experimental setup}
Throughout our experiments, we used the Gutenberg corpus as provided by the \textsc{nltk} platform, out of which we removed older texts (King John's Bible and Shakespeare). Sentences are enumerated two  by  two; each pair of sentences is then used as a distinct input source for \textsc{bert}. As we treat the \textsc{bert} algorithm as a black box, we retrieve only the embeddings from the last layer, discarding all intermediary representations and attention weights. We used the \texttt{bert-large-uncased} model in all experiments\footnote{Measurements were conducted before the release of the \texttt{bert-large-uncased-whole-words} model.}; therefore most of our experiments are done on word-pieces.

To study the basic coherence of \textsc{bert}'s semantic space, we can consider types as clusters of tokens---i.e.\ specific instances of contextualized embeddings---and thus leverage the tools of cluster analysis. In particular, silhouette score is generally used to assess whether a specific observation $\vec{v}$ is well assigned to a given cluster $C_i$ drawn from a set of possible clusters $C$. The silhouette score is defined in eq.\ \ref{eq:silh}:
\begin{align}
        \textit{sep}(\vec{v}, C_i) =& \min \{ \underset{\vec{v'} \in C_j}{\text{mean}} ~  d(\vec{v}, \vec{v'}) \forall ~ C_j  \in C - \{C_i\}  \}  \nonumber \\
        \textit{coh}(\vec{v}, C_i) =&  \underset{\vec{v'} \in C_i - \{\vec{v}\}}{\text{mean}} ~ d(\vec{v}, \vec{v'}) \nonumber \\
        \textit{silh}(\vec{v}, C_i) =& \frac{\textit{sep}(\vec{v}, C_i) -\textit{coh}(\vec{v}, C_i) }{ \max \{\textit{sep}(\vec{v}, C_i) , ~ \textit{coh}(\vec{v}, C_i) \}}  \label{eq:silh}
\end{align}
We used Euclidean distance for $d$. In our case, observations $\vec{v}$ therefore correspond to tokens (that is, \textsl{word-piece} tokens), and clusters $C_i$ to types. 

Silhouette scores consist in computing for each vector observation $\vec{v}$ a cohesion score (viz.\ the average distance to other observations in the cluster $C_i$) and a separation score (viz.\ the minimal average distance to other observations, i.e.\ the minimal `cost' of assigning $\vec{v}$ to any other cluster than $C_i$). Optimally, cohesion is to be minimized and separation is to be maximized, and this is reflected in the silhouette score itself: scores are defined between -1 and 1; -1 denotes that the observation $\vec{v}$ should be assigned to another cluster than $C_i$, whereas 1 denotes that the observation $\vec{v}$ is entirely consistent with the cluster $C_i$.
Keeping track of silhouette scores for a large number of vectors quickly becomes intractable, hence we use a slightly modified version of the above definition, and compute separation and cohesion using the distance to the average vector for a cluster rather than the average distance to other vectors in a cluster, as suggested by \citet{Vendramin2013silh}. Though results are not entirely equivalent as they ignore the inner structure of clusters, they still present a gross view of the consistency of the vector space under study.

We do note two caveats with our proposed methodology. Firstly, \textsc{bert} uses subword representations, and thus \textsc{bert} tokens do not necessarily correspond to words. However we may conjecture that some subwords exhibit coherent meanings, based on whether they tightly correspond to morphemes---e.g. `\textit{\#\#s}', `\textit{\#\#ing}' or `\textit{\#\#ness}'. %Moreover previous studies have shown that the larger part of the tokens generally correspond to words rather than subwords.
Secondly, we group word types based on character strings; yet only monosemous words should describe perfectly coherent clusters---whereas we expect some degree of variation for polysemous words and homonyms according to how widely their meanings may vary.

\subsection{Results \& Discussion}

\begin{figure}[t]
    \centering
    \includegraphics[scale=0.4]{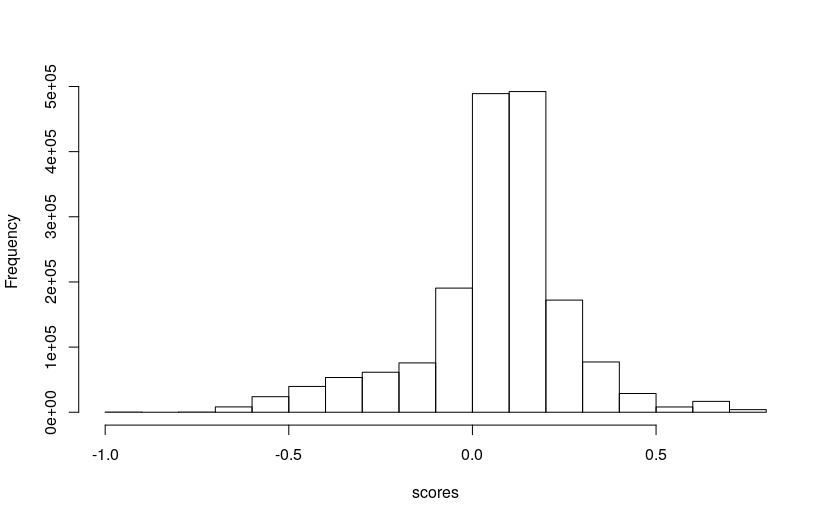}
    \caption{Distribution of token silhouette scores}
    \label{fig:vis-silhouette}
\end{figure}

\begin{comment}
\begin{table}[t]
    \centering
        \begin{tabular}{c c c c}
        \textbf{t-stat.}  & \textbf{$p$-val.} & \textbf{Cohen's $d$}  \\
        \hline
        -149.71 %df:& 1741600 
        & $< 2 \cdot 2^{-16}$ & -0.1208447
    \end{tabular}
    \caption{Paired t-test result for cohesion vs. separation}
    \label{tab:t-test-coh-vs-sep}
\end{table}
\end{comment}

We compared cohesion to separation scores using a paired Student's t-test, and found a significant effect ($p$-value $< 2 \cdot 2^{-16}$). %To estimate the effect size, we computed that Cohen's $d=-0.12084$ . %A visual representation of silhouette scores is also given in figure \ref{fig:vis-silhouette}.
%
%In short, the effect %---as shown by the t-test results---
This highlights that cohesion scores are lower than separation scores. The effect size as measured by Cohen's $d$ (Cohen's $d=-0.121$) is however rather small, suggesting that cohesion scores are only 12\% lower than separation scores. More problematically, we can see in figure \ref{fig:vis-silhouette} that 25.9\% of the tokens have a negative silhouette score: one out of four tokens would be better assigned to some other type than the one they belong to. %as can be seen in , not only do 17\% of the types have a negative median score---and thus contain at best as many well-clustered embeddings as misclassified ones---,
When aggregating scores by types, we found that 10\% of types contained only tokens with negative silhouette score.

%Though the overall picture seems satisfying, details show that t
The standards we expect of \textsc{dsm}s are not always upheld strictly; the median and mean score are respectively at 0.08 and 0.06, indicating a general trend of low scores, even when they are positive. %Some of our methodological choices may partly explain these: as we studied a pre-trained model, we computed statistics over subwords rather than words, and conjectured that the meaning representations associated to subwords ought to be similar to meaning representations associated to words. Likewise for the simplicity of our experiments, we ignored polysemy as we could not treat it consistently due to the non-standard tokenization. %Note however that a \textsc{glmm} to predict mean pairwise cosine per word type showed that number of meanings is a significant predictor, but does not account for the full cosine variation in and of itself.
We previously noted that both the use of sub-word representations %rather than word representations 
in \textsc{bert} as well as polysemy and homonymy might impact these results. The amount of meaning variation induced by polysemy and homonymy can be estimated by using a dictionary as a sense inventory. The number of distinct entries for a type serves as a proxy measure of how much its meaning varies in use.
We thus used a linear model to predict silhouette scores with log-scaled frequency and log-scaled definition counts, as listed in the Wiktionary, as predictors. We selected tokens for which we found at least one entry in the Wiktionary, out of which we then randomly sampled $10000$ observations. Both definition counts and frequency were found to be significant predictors, leading the silhouette score to decrease. This suggests that polysemy degrades the cohesion score of the type cluster, %reinforcing the view that \textsc{bert} representations cluster according to their types in a manner 
which is compatible with what one would expect from a \textsc{dsm}. We moreover observed that monosemous words yielded higher silhouette scores than polysemous words ($p < 2 \cdot 2^{-16}$, Cohen's $d = 0.236$), though they still include a substantial number of tokens with negative silhouette scores. %TODO: ADD TABLE

Similarity also includes related words, and not only tokens of the same type. Other studies \citep[e.g.]{vial19SenseVocComp,Coenen19BERTGeometry} already stressed that \textsc{bert} embeddings perform well on word-level semantic tasks. To directly assess whether \textsc{bert} captures this broader notion of similarity, we used the \textsc{men} word similarity dataset \citep{Bruni2014MEN}, which lists pairs of English words with human annotated similarity ratings. We removed pairs containing words for which we had no representation, leaving us with 2290 pairs. We then computed the Spearman correlation between similarity ratings and the cosine of the average \textsc{bert} embeddings of the two paired word types, and found a correlation of $0.705$, showing that cosine similarity of average \textsc{bert} embeddings encodes semantic similarity. For comparison, a word2vec \textsc{dsm} \citep[henceforth \textsc{w2v}]{Mikolov2013} trained on BooksCorpus \citep{Zhu2015BooksCorpus} using the same tokenization as \textsc{bert} achieved a correlation of $0.669$.

%\section{Segment encoding vectors} 
\section{Experiment 2: Cross-Sentence Coherence}
\label{sec:cross-sentence-coherence-per-type}
As observed in the previous section, overall the word type coherence in \textsc{bert} tends to match our basic expectations.
In this section, we do further tests, leveraging our knowledge of the design of \textsc{bert}. We detail the effects of jointly using \textsl{segment encodings} to distinguish between paired input sentences and \textsl{residual connections}.% and discuss whether these effects are not compatible with a distributional semantics interpretation of \textsc{bert} contextualized embeddings.
%
%In particular, we discuss the effects of using a \textit{segment encoding} to distinguish between sentences used in pairs as inputs jointly with the ubiquitous employment of residual connections throughout the model. %We first describe the issue at hand more formally before laying out our methodological and experimental protocol.

\subsection{Formal approach}
We begin by examining the architectural design of \textsc{bert}. We give some elements relevant to our study here and refer the reader to the original papers by \citet{Vaswani17} and \citet{Devlin18Bert}, introducing Transformers and \textsc{bert}, for a more complete description.
On a formal level, \textsc{bert} is a deep neural network composed of superposed layers of computations. Each layer is composed of two ``sub-layers'': the first performing ``multi-head attention'', the second being a simple feed-forward network. Throughout all layers, after each sub-layer, residual connections and layer normalization are applied, thus the intermediary output $\vec{o_L}$ after sub-layer $L$ can be written as a function of the input $\vec{x_L}$, as $\vec{o_L} = \textit{LayerNorm}(\textit{Sub}_L(\vec{x_L}) + \vec{x_L})$.

%As most neural networks, 
\textsc{bert} is optimized on two training objectives. The first, called \textsl{masked language model}, is a variation on the Cloze test for reading proficiency \citep{Taylor53Cloze}. 
The second, called \textsl{next sentence prediction} (\textsc{nsp}), corresponds to predicting whether two sentences are found one next to the other in the original corpus or not.
Each example passed as input to \textsc{bert} is comprised of two sentences, either contiguous sentences from a document, or randomly selected sentences. %this allows the model to provide an assessment of how well the model performs on this \textsc{nsp} objective for each input. For each example, a 
A special token \texttt{[SEP]} is used to indicate sentence boundaries, and the full sentence is prepended with a second special token \texttt{[CLS]} used to perform the actual prediction for \textsc{nsp}. Each token is transformed into an input vector using an input embedding matrix. % that maps word types to real-valued vector. 
To distinguish between tokens from the first and the second sentence, the model adds a learned feature vector $\vec{\textcolor{theblue}{\texttt{seg}_A}}$ to all tokens from first sentences, and a distinct learned feature vector $\vec{\textcolor{red}{\texttt{seg}_B}}$ to all tokens from second sentences; these feature vectors are called `segment encodings'. Lastly, as Transformer models %---unlike recurrent networks such as \textsc{lstm}s---
do not have an implicit representation of word-order, information regarding the index $i$ of the token in the sentence is added using a positional encoding $p(i)$. Therefore, if the initial training example was ``\textit{My dog barks. It is a pooch.}'', the actual input would correspond to the following sequence of vectors: 
\begin{align*}
         \vec{\texttt{[CLS]}} + \vec{p(0)} + \vec{\textcolor{theblue}{\texttt{seg}_A}}, & ~ \vec{My} + \vec{p(1)} + \vec{\textcolor{theblue}{\texttt{seg}_A}}, \\
         \vec{dog} + \vec{p(2)} + \vec{\textcolor{theblue}{\texttt{seg}_A}}, & ~ \vec{barks} + \vec{p(3)} + \vec{\textcolor{theblue}{\texttt{seg}_A}}, \\
         \vec{.} + \vec{p(4)} + \vec{\textcolor{theblue}{\texttt{seg}_A}}, & ~ \vec{\texttt{[SEP]}} + \vec{p(5)} + \vec{\textcolor{theblue}{\texttt{seg}_A}}, \\
         \vec{It} + \vec{p(6)} + \vec{\textcolor{red}{\texttt{seg}_B}}, & ~ \vec{is} + \vec{p(7)} + \vec{\textcolor{red}{\texttt{seg}_B}}, \\
         \vec{a} + \vec{p(8)} + \vec{\textcolor{red}{\texttt{seg}_B}}, & ~ \vec{pooch} + \vec{p(9)} + \vec{\textcolor{red}{\texttt{seg}_B}}, \\
         \vec{.} + \vec{p(10)} + \vec{\textcolor{red}{\texttt{seg}_B}}, & ~ \vec{\texttt{[SEP]}} + \vec{p(11)} + \vec{\textcolor{red}{\texttt{seg}_B}}
\end{align*}

Due to the general use of residual connections, marking the sentences using the segment encodings $\vec{\texttt{seg}_A}$ and $\vec{\texttt{seg}_B}$ can introduce a systematic offset within sentences.
Consider that the first layer uses as input vectors corresponding to word, position, and sentence information: $\vec{w_i} + \vec{p(i)} +\vec{\texttt{seg}_i}$; for simplicity, let $\vec{i_i}=\vec{w_i} + \vec{p(i)}$; we also ignore the rest of the input as it does not impact this reformulation. %As each sub-layer is applied a residual connection before normalization, t
The output from the first sub-layer $\vec{o^1_i}$ can be written:
\begin{align}
    \vec{o^1_i} &= \textit{LayerNorm}(\textit{Sub}_1(\vec{i_i} +\vec{\texttt{seg}_i}) + \vec{i_i} +\vec{\texttt{seg}_i}) \nonumber \\
            &= \vec{b}_l + \vec{g}^1 \odot \frac{1}{\sigma_i^1}\textit{Sub}_1(\vec{i_i}+\vec{\texttt{seg}_i} ) + \vec{g}^1 \odot \frac{1}{\sigma_i^1}  \vec{i_i}   \nonumber \\ 
            & \quad - \vec{g}^1 \odot \frac{1}{\sigma_i^1} \mu ({\textit{Sub}_1(\vec{i_i} +\vec{\texttt{seg}_i}) + \vec{i_i} +\vec{\texttt{seg}_i}})  \nonumber \\
            & \quad +\vec{g}^1 \odot\frac{1}{\sigma_i^1}  \vec{\texttt{seg}_i} \nonumber \\
            &  = \vec{\Tilde{o}}^1_i + \vec{g}^1 \odot \frac{1}{\sigma_i^1} \vec{\texttt{seg}_i}
\end{align}
This equation is obtain by simply injecting the definition for layer-normalization.\footnote{Layer normalization after sub-layer $l$ is defined as:
\begin{align*}
    \textit{LayerNorm}_l(\vec{x}) &= \vec{b}_l + \frac{\vec{g}_l \odot (\vec{x} - \mu(\vec{x}))}{\sigma} \\
        &=\vec{b}_l + \vec{g}_l \odot \frac{1}{\sigma} \vec{x} - \vec{g}_l \odot \frac{1}{\sigma} \mu(\vec{x}) 
\end{align*}
where $\vec{b_l}$ is a bias, $\odot$ denotes element-wise multiplication, $\vec{g_l}$ is a ``gain'' parameter,  $\sigma$ is the standard deviation of components of $\vec{x}$ and $\mu(\vec{x}) = \langle \bar{x}, \dots, \bar{x} \rangle$ is a vector with all components defined as the mean of components of $\vec{x}$.}
Therefore, by recurrence, the final output $\vec{o^L_i}$ for a given token $\vec{w_i} + \vec{p(i)} +\vec{\texttt{seg}_i}$ can be written as:
\begin{equation}
    \vec{o^L_i} = \vec{\Tilde{o}}^L_i +  \left( \bigodot \limits_{l=1}^L \vec{g}^l \right) \odot \left( \prod \limits_{l=1}^L \frac{1}{\sigma_i^l} \right) \times \vec{\texttt{seg}_i}
\end{equation}
%where $\vec{g'} = \bigodot \limits_{l=1}^L \vec{g}^l$. 
%with $n_i^1 = \textit{LayerNorm}(\textit{Sub}_1(\vec{i_i} +\vec{\texttt{seg}_i}) + \vec{i_i} +\vec{\texttt{seg}_i})$.

%\begin{comment}
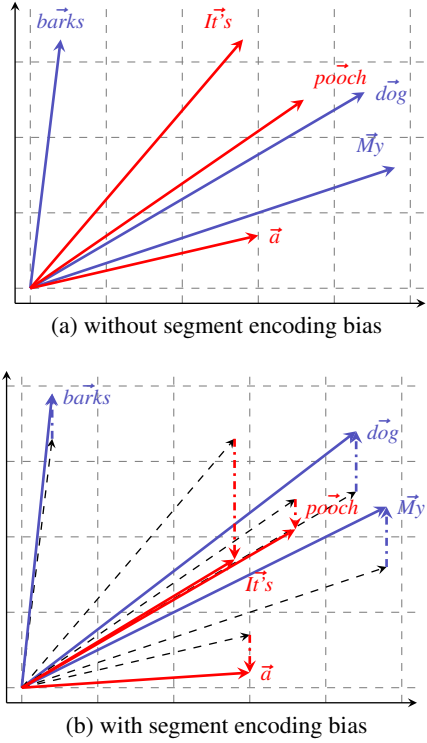
\begin{figure}[t]
    \centering
    \subfloat[without segment encoding bias]{
        \begin{tikzpicture}
			\draw[dashed, gray](-0.2, 0) -- (5.2,0) ;
			\draw[dashed, gray](-0.2, 1) -- (5.2,1) ;
			\draw[dashed, gray](-0.2, 2) -- (5.2,2) ;
			\draw[dashed, gray](-0.2, 3) -- (5.2,3) ;
			%\draw[dashed, gray](-0.2, 4) -- (5.2,4) ;
			%\draw[dashed, gray](-0.2, 5) -- (5.2,5) ;
			\draw[dashed, gray](0, -0.2) -- (0,3.8) ;
			\draw[dashed, gray](1, -0.2) -- (1,3.8) ;
			\draw[dashed, gray](2, -0.2) -- (2,3.8) ;
			\draw[dashed, gray](3, -0.2) -- (3,3.8) ;
			\draw[dashed, gray](4, -0.2) -- (4,3.8) ;
			\draw[dashed, gray](5, -0.2) -- (5,3.8) ;
			\draw[line width=0.5pt, -stealth](-0.2,-0.2)->(-0.2,3.8);
			\draw[line width=0.5pt, -stealth](-0.2,-0.2)->(5.2,-0.2);
    		\tkzDefPoint(4.4,2.6){np1}
    		\tkzDefPoint(4.8,1.6){np2}
    		\tkzDefPoint(0.4,3.3){nq1}
    		
    		\tkzDefPoint(4.4,3.4){np1s}
    		\tkzDefPoint(4.8,2.4){np2s}
    		\tkzDefPoint(0.4,3.9){nq1s}
    		
    		\tkzDefPoint(3.6,2.5){nq2}
    		\tkzDefPoint(2.8, 3.3){npp}
    		\tkzDefPoint(3, 0.7){nqp}
    		
    		\tkzDefPoint(3.6,2.1){nq2s}
    		\tkzDefPoint(2.8, 1.7){npps}
    		\tkzDefPoint(3, 0.2){nqps}
    		
    		\tkzDefPoint(0, 0){o}

    		\draw[line width=1pt, theblue, -stealth]( o)--(np1) node[pos=1, right]{$\scriptstyle\boldsymbol{\vec{\textit{dog}}}$};
    		\draw[line width=1pt, theblue, -stealth]( o)--(np2) node[pos=1,above left]{$\scriptstyle\boldsymbol{\vec{\textit{My}}}$};
    		\draw[line width=1pt, theblue, -stealth]( o)--(nq1) node[pos=1,above]{$\scriptstyle\boldsymbol{\vec{\textit{barks}}}$};
    		
    		%\draw[line width=1pt, dash dot,  red, -stealth](np1)--(np1s) ;
    		%\draw[line width=1pt, dash dot, red, -stealth](np2)--(np2s) ;
    		%\draw[line width=1pt, dash dot, red, -stealth](nq1)--(nq1s) ;
    		
    		%\draw[line width=1pt, red, -stealth](o)--(np1s) ;
    		%\draw[line width=1pt, red, -stealth](o)--(np2s) ;
    		%\draw[line width=1pt, red, -stealth](o)--(nq1s) ;
    
    		\draw[line width=1pt, red, -stealth]( o)--(npp) node[pos=1,above left]{$\scriptstyle\boldsymbol{\vec{\textit{It's}}}$};
    		\draw[line width=1pt, red, -stealth]( o)--(nq2) node[pos=1,above right]{$\scriptstyle\boldsymbol{\vec{\textit{pooch}}}$};
    		\draw[line width=1pt, red, -stealth](o)--(nqp) node[pos=1, right]{$\scriptstyle\boldsymbol{\vec{\textit{a}}}$};
    		
    		%\draw[line width=1pt, dash dot, theblue, -stealth]( npp)--(npps);
    		%\draw[line width=1pt, dash dot, theblue, -stealth]( nq2)--(nq2s);
    		%\draw[line width=1pt, dash dot, theblue, -stealth]( nqp)--(nqps);
    		
    		%\draw[line width=1, theblue, -stealth]( o)--(npps);
    		%\draw[line width=1pt, theblue, -stealth]( o)--(nq2s);
    		%\draw[line width=1pt, theblue, -stealth]( o)--(nqps);
        \end{tikzpicture}} \\
    \subfloat[with segment encoding bias]{
        \begin{tikzpicture}
			\draw[dashed, gray](-0.2, 0) -- (5.2,0) ;
			\draw[dashed, gray](-0.2, 1) -- (5.2,1) ;
			\draw[dashed, gray](-0.2, 2) -- (5.2,2) ;
			\draw[dashed, gray](-0.2, 3) -- (5.2,3) ;
			\draw[dashed, gray](-0.2, 4) -- (5.2,4) ;
			%\draw[dashed, gray](-0.2, 5) -- (5.2,5) ;
			\draw[dashed, gray](0, -0.2) -- (0,4.2) ;
			\draw[dashed, gray](1, -0.2) -- (1,4.2) ;
			\draw[dashed, gray](2, -0.2) -- (2,4.2) ;
			\draw[dashed, gray](3, -0.2) -- (3,4.2) ;
			\draw[dashed, gray](4, -0.2) -- (4,4.2) ;
			\draw[dashed, gray](5, -0.2) -- (5,4.2) ;
			\draw[line width=0.5pt, -stealth](-0.2,-0.2)->(-0.2,4.2);
			\draw[line width=0.5pt, -stealth](-0.2,-0.2)->(5.2,-0.2);
    		\tkzDefPoint(4.4,2.6){np1}
    		\tkzDefPoint(4.8,1.6){np2}
    		\tkzDefPoint(0.4,3.3){nq1}
    		
    		\tkzDefPoint(4.4,3.4){np1s}
    		\tkzDefPoint(4.8,2.4){np2s}
    		\tkzDefPoint(0.4,3.9){nq1s}
    		
    		\tkzDefPoint(3.6,2.5){nq2}
    		\tkzDefPoint(2.8, 3.3){npp}
    		\tkzDefPoint(3, 0.7){nqp}
    		
    		\tkzDefPoint(3.6,2.1){nq2s}
    		\tkzDefPoint(2.8, 1.7){npps}
    		\tkzDefPoint(3, 0.2){nqps}
    		
    		\tkzDefPoint(0, 0){o}
    		\draw[line width=0.5pt, dashed , -stealth]( o)--(np1) ;
    		\draw[line width=0.5pt, dashed , -stealth]( o)--(np2) ;
    		\draw[line width=0.5pt, dashed,  -stealth]( o)--(nq1) ;
    		
    		\draw[line width=1pt, dash dot,  theblue, -stealth](np1)--(np1s) ;
    		\draw[line width=1pt, dash dot, theblue, -stealth](np2)--(np2s) ;
    		\draw[line width=1pt, dash dot, theblue, -stealth](nq1)--(nq1s) ;
    		
    		\draw[line width=1pt, theblue, -stealth](o)--(np1s) node[pos=1, right]{$\scriptstyle\boldsymbol{\vec{\textit{dog}}}$};
    		\draw[line width=1pt, theblue, -stealth](o)--(np2s) node[pos=1, right]{$\scriptstyle\boldsymbol{\vec{\textit{My}}}$};
    		\draw[line width=1pt, theblue, -stealth](o)--(nq1s) node[pos=1,right]{$\scriptstyle\boldsymbol{\vec{\textit{barks}}}$};
    
    		\draw[line width=0.5pt, dashed,  -stealth]( o)--(npp) ;
    		\draw[line width=0.5pt, dashed,   -stealth]( o)--(nq2) ;
    		\draw[line width=0.5pt, dashed,   -stealth](o)--(nqp) ;
    		
    		\draw[line width=1pt, dash dot, red, -stealth]( npp)--(npps);
    		\draw[line width=1pt, dash dot, red, -stealth]( nq2)--(nq2s);
    		\draw[line width=1pt, dash dot, red, -stealth]( nqp)--(nqps);
    		
    		\draw[line width=1, red, -stealth]( o)--(npps) node[pos=1,below right]{$\scriptstyle\boldsymbol{\vec{\textit{It's}}}$};
    		\draw[line width=1pt, red, -stealth]( o)--(nq2s) node[pos=1,above right]{$\scriptstyle\boldsymbol{\vec{\textit{pooch}}}$};
    		\draw[line width=1pt, red, -stealth]( o)--(nqps) node[pos=1, right]{$\scriptstyle\boldsymbol{\vec{\textit{a}}}$};
        \end{tikzpicture}}
    \caption{Segment encoding bias}
    \label{fig:segment-encoding-bias}
\end{figure}
%\end{comment}
%

%
This rewriting trick shows that segment encodings are partially preserved in the output. All embeddings within a sentence contain a shift in a specific direction, determined only by the initial segment encoding and the learned gain parameters for layer normalization. %In principle, this shift may be negligible compared to the remaining part of the output $\vec{\Tilde{o}}^L_i$. %MC: changed for below
%In principle, this segment encoding bias may be negligible in the output embedding.
%In the terms of distributional semantics, the semantic space of tokens in second sentences is possibly shifted with respect to the semantic space of tokens in first sentences.
In figure \ref{fig:segment-encoding-bias}, we illustrate what this systematic shift might entail. Prior to the application of the segment encoding bias, the semantic space is structured by similarity (`\textit{pooch}' is near `\textit{dog}'); with the bias, we find a different set of characteristics: in our toy example, tokens are linearly separable by sentences.

\subsection{Data \& Experimental setup}

If \textsc{bert} properly describes a  semantic vector space,  %, and not influenced by segment encodings. 
we should, on average, observe no significant difference in token encoding imputable to the segment the token belongs to. For a given word type $w$, we may constitute two groups: $w_{\texttt{seg}_A}$, the set of tokens for this type $w$ belonging to first sentences in the inputs, and $w_{\texttt{seg}_B}$, the set of tokens of $w$ belonging to second sentences. If \textsc{bert} counterbalances the segment encodings, %and no effect can be perceived,
random differences should cancel out, and therefore the mean of all tokens $w_{\texttt{seg}_A}$ should be equivalent to the mean of all tokens $w_{\texttt{seg}_B}$.

We used the same dataset as in section \ref{sec:word-type-cohesion}. This setting (where all paired input sentences are drawn from running text) %, using running sentences %ensures us that all second sentences in the input followed the first in the original text, thus neutralizes any other possible impact of the \textsc{nsp} objective, and 
allows us to focus on the effects of the segment encodings. % themselves. %
We retrieved the output embeddings of the last \textsc{bert} layer and grouped them per word type. % and segment encodings---viz. we constituted distinct groups for word $w$ attested in first sentences and the same word $w$ attested in second sentences.
To assess the consistency of a group of embeddings with respect to a purported reference, we used a mean of squared error (\textsc{mse}): given a group of embeddings $E$ and a reference vector $\vec{r}$, we computed how much each vector in $E$ strayed from the reference $\vec{r}$. It is formally defined as:
\begin{equation}\label{eq:mse}
    \text{MSE}(E, \vec{r}) = \frac{1}{\# E} \sum_{\vec{v} \in E} \sum_d (\vec{v}_d - \vec{r}_d)^2
\end{equation}
This \textsc{mse} can also be understood as the average squared distance to the reference $\vec{r}$. When $\vec{r}=\overline{E}$, i.e.\ $\vec{r}$ is set to be the average vector in $E$, the \textsc{mse} measures variance of $E$ via Euclidean distance.
We then used the \textsc{mse} function to construct pairs of observations: for each word type $w$, and for each segment encoding $\texttt{seg}_i$, we computed two scores: $\text{MSE}(w_{\texttt{seg}_i}, \overline{w_{\texttt{seg}_i}})$---which gives us an assessment of how coherent the set of embeddings $w_{\texttt{seg}_i}$ is with respect to the mean vector in that set---and $\text{MSE}(w_{\texttt{seg}_i}, \overline{w_{\texttt{seg}_j}})$---which assesses how coherent the same group of embeddings is with respect to the mean vector for the embeddings of the same type, but from the other segment $\texttt{seg}_j$. If no significant contrast between these two scores can be observed, then \textsc{bert} counterbalances the segment encodings and is coherent across sentences.

\subsection{Results \& Discussion}

\begin{figure}[ht]
    \centering
    \includegraphics[scale=0.5]{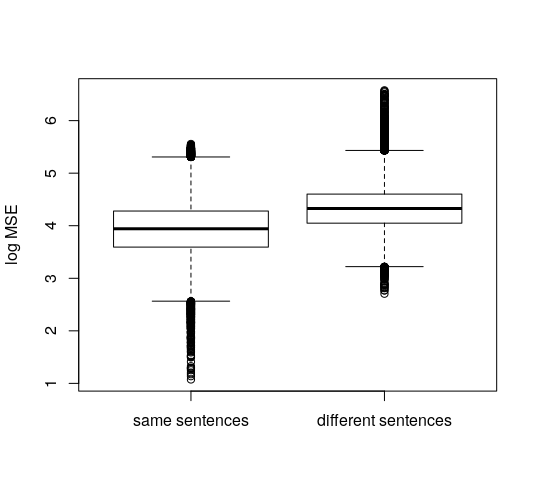}
    \caption{Log-scaled \textsc{mse} per reference}% TODO: ADD PROPER CAPTIONS FOR X AND Y AXES. THE R CODE THAT IS THERE NOW IS NOT READER-FRIENDLY}
    \label{fig:log-scaled-mse-type-cohesion}
\end{figure}

\begin{comment}
\begin{table}[th]
    \centering
    \begin{tabular}{c c c}
        \textbf{t-stat.} %& \textbf{df.} 
        & \textbf{$p$-val.} & \textbf{Cohen's $d$}  \\
        \hline
         -117.95 %& 25144 
         & $< 2\cdot 2^{-16}$ & -0.5268134 \\
    \end{tabular}
    \caption{Paired t-test results between \textsc{mse} scores}
    \label{tab:t-test-sentence-drift}
\end{table}
\end{comment}

We compared results using a paired Student's t-test, which highlighted a significant difference based on which segment types belonged to ($p$-value $<2\cdot 2^{-16}$); 
%(cf. table \ref{tab:t-test-sentence-drift}); 
the effect size (Cohen's $d=-0.527$) was found to be stronger than what we computed when assessing whether tokens cluster according to their types (cf. section \ref{sec:word-type-cohesion}). A visual representation of these results, log-scaled, is shown in figure \ref{fig:log-scaled-mse-type-cohesion}. For all sets $w_{\texttt{seg}_i}$, the average embedding from the set itself was systematically a better fit than the average embedding from the paired set $w_{\texttt{seg}_j}$. We also noted that a small number of items yielded a disproportionate difference in \textsc{mse} scores %Another trend we observed was that frequency played a role.%, as can be seen in figure \ref{fig:log-log-freq-gap}, though it did not subsume the entire distribution: 
%We also observed
and that frequent word types had smaller differences in \textsc{mse} scores: roughly speaking, very frequent items---punctuation signs, stop-words, frequent word suffixes---received embeddings that are \textsl{almost} coherent across sentences.
%We however note that many downstream applications use a single segment encoding  per input, presumably avoiding the caveat stressed here.

Although the observed positional effect of embeddings' inconsistency might be entirely due to segment encodings, additional factors might be at play.
In particular, \textsc{bert} uses absolute positional encoding vectors to order words within a sequence: the first word $w_1$ is marked with the positional encoding $p(1)$, the second word $w_2$ with $p(1)$, and so on until the last word, $w_n$, marked with $p(n)$. As these positional encodings are added to the word embeddings, the same remark made earlier on the impact of residual connections may apply to these positional encodings as well. Lastly, we also note that many downstream applications use a single segment encoding per input, and thus sidestep the caveat stressed here.

\section{Experiment 3: Sentence-level structure} \label{sec:lexical-contrasts-and-cross-sentence-coherence}

We have seen %in section \ref{sec:cross-sentence-coherence-per-type} 
that \textsc{bert} embeddings do not fully respect cross-sentence coherence; the same type receives somewhat different representations for occurrences in even and odd sentences. However, comparing tokens of the same type in consecutive sentences is not necessarily the main application of \textsc{bert} and related models. Does the segment-based representational variance affect the structure of the semantic space, instantiated in similarities between tokens of different types? %. might Perhaps %A related question is whether segment encodings have an effect outside of the shift we saw for tokens: they could also 
Here we investigate how segment encodings impact the relation between any two tokens in a given sentence. %Moreover, studying paired items inside a sentence gives us an insight on how \textsc{bert} structures sentences, and whether that treatment is consistent across sentences positions.

\subsection{Data \& Experimental setup}

Consistent with previous experiments, we used the same dataset (cf. section \ref{sec:word-type-cohesion}); in this experiment also mitigating the impact of the \textsc{nsp} objective was crucial. Sentences were thus passed two by two as input to the \textsc{bert} model. %; all outputs were concatenated so as to obtain observations per tokens.
As cosine has been traditionally used to quantify semantic similarity between words \citep[e.g.]{Mikolov2013a,Levy2014a}, we then computed pairwise cosine of the tokens in each sentence. This allows us to reframe our assessment of whether lexical contrasts are coherent across sentences as a comparison of semantic dissimilarity across sentences. More formally, we compute the following set of cosine scores $C_S$ for each sentence $S$:
\begin{equation}
    C_S = \{ \cos (\vec{v}, \vec{u}) ~ | ~ \vec{v} \neq \vec{u} ~ \land ~ \vec{v}, \vec{u} \in E_S \} 
\end{equation}
with $E_S$ the set of embeddings for the sentence $S$. In this analysis, we compare the union of all sets of cosine scores for first sentences against the union of all sets of cosine scores for second sentences.
To avoid asymmetry, we remove the \texttt{[CLS]} token (only present in first sentences), and as with previous experiments we neutralize the effects of the \textsc{nsp} objective by using only consecutive sentences as input.

\subsection{Results \& Discussion}

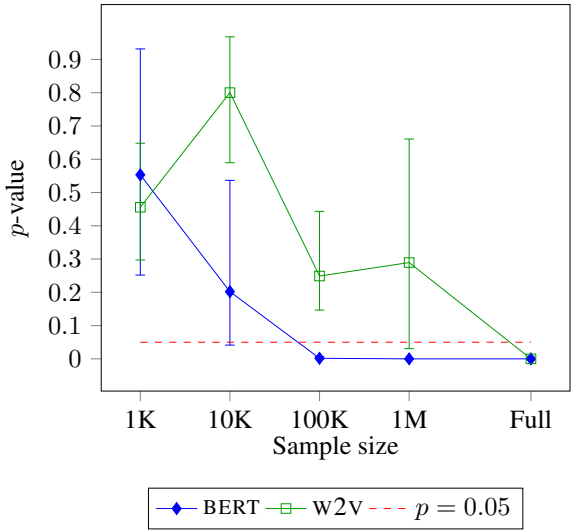
\begin{figure}[ht]
    \centering
    \begin{tikzpicture}[scale=0.9]
\begin{axis}[
  xmode=log,
  log ticks with fixed point,
  ytick={0,0.1,...,1}, ytick align=outside, ytick pos=left,
  xtick={1000,10000,100000,1000000, 22869901}, xtick align=outside, xtick pos=left, xticklabels={1K,10K,100K, 1M, Full},
  xlabel={Sample size},
  ylabel={$p$-value},
  legend style={at={(0.5,-.25)},anchor=north, legend columns=4,}]
\addplot+[
  blue, mark=diamond*, mark options={blue, scale=1.25},
  error bars/.cd, 
    y dir=both, 
    y explicit
]
table [x=x, y=y, y error plus=ymax, y error minus=ymin, col sep=comma] {
    x,  error,       y,    ymax,  ymin
    1000,  0.3448312,  0.55326,    0.37854,  0.30136
    10000,  0.2017489,  0.20202,   0.33458,  0.16069
    100000,  0.0028226684,  0.001653098,   0.004875902,   0.0016530569
    1000000,  0,  0,    0,  0
    22869901, 0, 0,    0,  0
};
\addlegendentry{\textsc{bert}};

\addplot+[
  green!60!black, mark=square, mark options={green!60!black, scale=1},
  error bars/.cd, 
    y fixed,
    y dir=both, 
    y explicit
] table [x=x, y=y,y error plus=ymax, y error minus=ymin,, col sep=comma] {
    x,  error,       y, ymax,   ymin
    1000,  0.1478518,  0.45578,  0.19252,   0.15858
    10000,  0.1665097,  0.80032,    0.16788,    0.21062
    100000,  0.1243906,  0.24894,   0.19446,     0.10234
    1000000,  0.2537503,  0.28953,  0.37127,     0.25878
    22869901, 0, 8.732e-11,    0,  0
    
};
\addlegendentry{\textsc{w2v}};
\addplot+[mark=none, red, dashed, samples=2, domain=1000:22869901] {0.05};
\addlegendentry{$p=0.05$};
\end{axis}
\end{tikzpicture}
    \caption{Wilcoxon tests, 1\textsuperscript{st} vs. 2\textsuperscript{nd} sentences}
    \label{fig:p-val-as-sampfunc}
\end{figure}

We compared cosine scores for first and second sentences using a Wilcoxon rank sum test. %; we did not use a Welch $t$-test as the data was not normally distributed. %Figure \ref{fig:pairwise-cos-sentence-level} also shows a visual display of distribution of cosine scores.
We observed a significant effect, however small (Cohen's $d = 0.011$). This may perhaps be due to data idiosyncrasies, and indeed when comparing with a \textsc{w2v} \citep{Mikolov2013} trained on BooksCorpus \citep{Zhu2015BooksCorpus} using the same tokenization as \textsc{bert}, we do observe a significant effect ($p<0.05$). However the effect size is six times smaller ($d=0.002$) %($d=0.00184$)
than what we found for \textsc{bert} representations; moreover, when varying the sample size (cf. figure \ref{fig:p-val-as-sampfunc}), $p$-values for \textsc{bert} representations drop much faster to statistical significance. 

A possible reason for the larger discrepancy observed in \textsc{bert} representations might be that \textsc{bert} uses absolute positional encodings, i.e. the k\textsuperscript{th} word of the input is encoded with $p(k)$. Therefore, although all first sentences of a given length $l$ will be indexed with the same set of positional encodings $\{p(1), ~ \dots, ~ p(l)\}$, only second sentences of a given length $l$ preceded by first sentences of a given length $j$ share the exact same set of positional encodings $\{p(j+1), ~ \dots, ~ p(j+l)\}$. As highlighted previously, the residual connections ensure that the segment encodings were partially preserved in the output embedding: the same argument can be made for positional encodings. %, with the caveat that only segment encodings are applied to all words in a given sentence.
In any event, the fact is that we do observe on \textsc{bert} representations an effect of segment on sentence-level structure. This effect is greater than one can blame on data idiosyncrasies, as verified by the comparison with a traditional \textsc{dsm} such as \textsc{w2v}. If we are to consider \textsc{bert} as a \textsc{dsm}, we must do so at the cost of cross-sentence coherence.%: although it may be useful in some settings such as discourse analysis or computational semantics, one may question the pertinence of such a feature if \textsc{dsm}s are to purely model lexical semantics.

The %violation of cross-sentence coherence
analysis above suggests that embeddings for tokens drawn from first sentences live in a different semantic space than tokens drawn from second sentences, i.e.\ that \textsc{bert} contains two \textsc{dsm}s rather than one. If so, the comparison between two sentence-representations from a single input would be meaningless, or at least less coherent than the comparison of two sentence representations drawn from the same sentence position.
To test this conjecture, we use two compositional semantics benchmarks: \textsc{sts} \citep{Cer17STS} and \textsc{sick-r} \citep{marelli-etal-2014-sick}. These datasets are structured as triplets, grouping a pair of sentences with a human-annotated relatedness score. The original presentation of \textsc{bert} \citep{Devlin18Bert} did include a downstream application to these datasets, but employed a learned classifier, which obfuscates results \citep{Wieting2019randsent,cover65,Hewitt2019ProbingProbes}. Hence we simply reduce the sequence of tokens within each sentence into a single vector by summing them, a simplistic yet robust semantic composition method. We then compute the Spearman correlation between the cosines of the two sum vectors and the sentence pair's relatedness score. We compare two setups: a ``two sentences input'' scheme (or \textit{2 sent. ipt.} for short)---where we use the sequences of vectors obtained by passing the two sentences as a single input---and a ``one sentence input'' scheme (\textit{1 sent. ipt.})---using two distinct inputs of a single sentence each.

\begin{table}[t]
    \centering
    \npdecimalsign{.}
    \nprounddigits{5}

\begin{tabular}{|l n{2}{5} n{2}{5}|}
    \hline
    \textbf{Model} & \textbf{\textsc{sts} cor.} & {{\textbf{\textsc{sick-r} cor.}}} %& p-value  
    \\ \hline \hline
    %average vector (\textsc{bow}) & 0.23843531478046912 %& 3.977169224704639e-75 
    %\\ \hline
    %component-wise multiplication  & -0.004410580829412534 & 0.7381164657791739 \\ \hline
    %first vector & 0.039194983662628025 %& 0.0029553516344784855  
    %\\ \hline \hline
    Skip-Thought & 0.2555976611661546  & 0.48761929800248277
    \\ %\hline
    \textsc{use} & 0.6668647269750902 & 0.6899666474448651
    \\ %\hline
    InferSent  & 0.6764647661724299  & 0.7090345316003841 %, pvalue=0.0)
    \\ \hline \hline 
    %Open\textsc{nmt} on EuroParl en $\rightarrow$ de & 0.15228299484345043 %& 3.5941482414675722e-31
    %\\ \hline 
    % Open\textsc{nmt} on EuroParl en $\rightarrow$ de + $\sum$ & 0.2032140008393688 %& 3.5941482414675722e-31
    %\\ \hline 
    %Open\textsc{nmt} on OpenSubtitles en $\rightarrow$ es  & 0.28080784398761927% & 1.1805025025979695e-104 
    %\\ \hline 
    % Open\textsc{nmt} on OpenSubtitles en $\rightarrow$ es + $\sum$  & 0.1981501085104484 \\ \hline 
    %\textsc{rvd} + cosine loss & 0.20258980118460743  %&2.5949610226732514e-54
    %\\ \hline 
    %\textsc{lstm} + \textsc{rvd} & 0.3663852284045215 % & 3.7344626380779554e-182
    %\\ \hline \hline
    \textsc{bert}, \textit{2 sent. ipt.} & 0.35912729311102576 & 0.36991548426408144   \\ %\hline
    \textsc{bert}, \textit{1 sent. ipt.}  & 0.4824134048301114  & 0.5869474034197179 \\
    \textsc{w2v}    & 0.3701692035923143  & 0.5335559401867137 \\ \hline
    %\texttt{[CLS]} token  &  & 0.4274507398921791  \\ \hline
    %\textsc{bert} + $\sum $ + \textsc{rvd} $\rightarrow$ \textsc{ft}   & 0.602666799174115 \\ \hline
    %\textsc{use}+ \textsc{rvd} $\rightarrow$ \textsc{use} & 0.5308208281209419
    %\\ \hline
    %Infersent + \textsc{rvd} $\rightarrow$ Infersent & 0.6446959570104261 \\ \hline

\end{tabular}
\caption{\label{tab:sts}Correlation (Spearman $\rho$) of cosine similarity and relatedness ratings on the \textsc{sts} and \textsc{sick-r} benchmarks}
\end{table}

Results are reported in table \ref{tab:sts}; we also provide comparisons with three different sentence encoders and the aforementioned \textsc{w2v} model.
%Comparing performances of the two setups seems to confirm our expectations: 
As we had suspected, using sum vectors drawn from a two sentence input scheme single degrades performances below the \textsc{w2v} baseline.
%These results are therefore not compatible with the distributional interpretation of cross-sentence incoherence that we previously suggested---namely, that \textsc{bert} embeddings capture some form of discourse-level information such that word representations from second sentences benefit from a richer introductory context. If it were the case, then we would expect this discourse-level information to be leveraged when using two sentences at once as input, and that such a setup would yield the best performances. 
On the other hand, a one sentence input scheme seems to produce coherent sentence representations: in that scenario, \textsc{bert} performs better than \textsc{w2v} and the older sentence encoder Skip-Thought \citep{SkipthoughtsKiros15}, but worse than the modern \textsc{use} \citep{Cer18USE} and Infersent \citep{InfersentConneau17}. The comparison with \textsc{w2v} also shows that \textsc{bert} representations over a coherent input are more likely to include some form of compositional knowledge than traditional \textsc{dsm}s; however it is difficult to decide whether some true form of compositionality is achieved by \textsc{bert} or whether these performances are entirely a by-product of the positional encodings. In favor of the former, other research has suggested that Transformer-based architectures perform syntactic operations \citep{raganato-tiedemann-2018-analysis,Hewitt2019ASP,Clark19BertLook,jawahar19BERTStructLang,voita19pruninghead,Michel19Sixteen}. %, although the methodology employed in these works has been met with criticism \citep{Brunner2019Validity,serrano2019Interpretable}.
%
%Some of the discrepancy in lexical contrast that we observed, along with the encouraging performances on compositionality benchmarks, may therefore be tentatively explained as the \textsc{bert} embedding model learning to contrast pairs of sentence. For the contrast to be significant, the model must have some form of understanding of what the objects to be contrasted are; this in turn requires some form of compositional knowledge to abstract away from mere sequences of tokens.
In all, these results suggest that the semantic space of token representations from second sentences differ from that of embeddings from first sentences.

\section{Conclusions} \label{sec:ccl}

%Write something about how we tested whether BERT token representations have the properties we expect from semantic vectors. 

Our experiments have focused on testing to what extent similar words lie in similar regions of \textsc{bert}'s latent semantic space. % and what are the consequences of the next-sentence prediction objective of \textsc{bert} on the vector representations (sections \ref{sec:cross-sentence-coherence-per-type}, \ref{sec:lexical-contrasts-and-cross-sentence-coherence})
Although we saw that type-level semantics seem to match our general expectations about \textsc{dsm}s, focusing on details leaves us with a much foggier picture.

The main issue stems from \textsc{bert}'s ``next sentence prediction objective'', which requires tokens to be marked according to which sentence they belong. This introduces a distinction between \textsl{first} and \textsl{second sentence of the input} that runs contrary to our expectations in terms of cross-sentence coherence. 
The validity of such a distinction for lexical semantics may be questioned, yet its effects can be measured. The primary assessment conducted in section \ref{sec:word-type-cohesion} shows that token representations did tend to cluster naturally according to their types, yet a finer study detailed in section \ref{sec:cross-sentence-coherence-per-type} highlights that tokens from distinct sentence positions (even vs.\ odd) tend to have different representations. This can seen as a direct consequence of \textsc{bert}'s architecture: residual connections, along with the use of specific vectors to encode sentence position, entail that tokens for a given sentence position are `shifted' with respect to tokens for the other position.
%
%We also show in section \ref{sec:lexical-contrasts-and-cross-sentence-coherence} that the use of two sentences as input introduces a variation in lexical contrasts above what can be blamed on corpus specificity. 
Encodings have a substantial effect on the structure of the semantic subspaces of the two sentences in \textsc{bert} input. Our experiments demonstrate that assuming sameness of the semantic space across the two input sentences can lead to a significant performance drop in semantic textual similarity.

One way to overcome this violation of cross-sentence coherence would be to consider first and second sentences representations as belonging to distinct distributional semantic spaces. %though we may find some form of evidence for semantic composition in \textsc{bert} representations, paired sentences that violate cross-sentence coherence do not improve performances on semantic composition benchmarks.
The fact %When focusing on a purely mechanistic account of these facts, 
that first sentences were shown to have on average higher pairwise cosines than second sentences can be partially explained by the use of absolute positional encodings in \textsc{bert} representations. Although positional encodings are required so that the model does not devolve into a bag-of-word system, absolute encodings are not: other works have proposed alternative relative position encodings %elsewhere in the literature
\citep[e.g.]{shaw18rel,dai19tfxl}; replacing the former with the latter may alleviate the gap in lexical contrasts. Other related questions that we must leave to future works encompass testing on other \textsc{bert} models such as the whole-words model, or that of \citet{liu19roberta} which differs only by its training objectives, as well as other contextual embeddings architectures. 

Our findings suggest that the formulation of the \textsc{nsp} objective of \textsc{bert} obfuscates its relation to distributional semantics, by introducing a systematic distinction between first and second sentences which impacts the output embeddings. Similarly, other works \citep{lample19xlingpretrain,xlnet,Joshi19SpanBert,liu19roberta} stress that the usefulness and pertinence of the \textsc{nsp} task were not obvious. %; likewise the other training objective of \textsc{bert} was shown to introduce an artificial distinction between training and application. %and that the other objective---the \textsl{masked language model} task---also introduced an artificial distinction between training and downstream application. 
These studies favored an empirical point of view; here, we have shown what sorts of caveats came along with such artificial distinctions from the perspective of a theory of lexical semantics. %Further research is due to compare these findings across corpora and embedding architectures, contextual or not. 
We hope that future research will extend and refine these findings, and further our understanding of the peculiarities of neural architectures as models of linguistic structure.

\section*{Acknowledgments}
We thank Quentin Gliosca whose remarks have been extremely helpful to this work. We also thank Olivier Bonami as well as three anonymous reviewers for their thoughtful criticism. 
The work was supported  by a public grant overseen by the French National Research Agency (ANR)  as part of the ``Investissements d'Avenir'' program: Idex \emph{Lorraine Universit\'e d'Excellence} (reference: ANR-15-IDEX-0004).

\bibliography{naaclhlt2018}
\bibliographystyle{acl_natbib}

\end{document}